%% file: PaperForReview.tex
\pdfoutput=1

\documentclass[11pt]{article}

\usepackage{EMNLP2022}

\usepackage{times}
\usepackage{latexsym}

\usepackage[ruled,vlined]{algorithm2e}
\usepackage{hyperref}       
\usepackage{url}            
\usepackage{booktabs}       
\usepackage{nicefrac}       
\usepackage{microtype}      
\usepackage{xcolor}         
\usepackage{graphicx}
\usepackage{float}
\usepackage{adjustbox}
\usepackage{array}
\usepackage{multirow}
\usepackage{subcaption}
\usepackage{placeins}
\usepackage{makecell}
\usepackage{newfloat}
\usepackage{listings}
\usepackage{bibentry}
\usepackage{wrapfig}
\usepackage{pifont}
\usepackage{arydshln}
\usepackage{amsmath,amssymb,amsfonts}
\usepackage[noend]{algpseudocode}
\usepackage{bm}	
\usepackage{bbm}

\usepackage[T1]{fontenc}
\usepackage[utf8]{inputenc}

\usepackage{microtype}
 \usepackage{mathrsfs}
\usepackage{inconsolata}
\usepackage{amsfonts}

\usepackage{tikz}       %
\usepackage{tcolorbox}
\usetikzlibrary{shapes.misc}
\usetikzlibrary{shadows}

\DeclareRobustCommand{\tcgray}[1]{
\begin{tikzpicture}[baseline=(char.base)]
\node(char)[
  draw,fill=black!15,
  shape=rounded rectangle,
  text height=5pt,
  drop shadow={opacity=.5,shadow xshift=0pt,shadow yshift=-1pt},
]
  {\normalfont #1};
\end{tikzpicture}
}
\DeclareRobustCommand{\tcblack}[1]{
\begin{tikzpicture}[baseline=(char.base)]
\node(char)[
  draw,fill=black,
  shape=rounded rectangle,
  text height=5pt,
  drop shadow={opacity=.5,shadow xshift=0pt,shadow yshift=-1pt},
]
  {\color{white}{\normalfont #1}};
\end{tikzpicture}
}

\title{$\mathcal{S}$parse$\mathcal{A}$dapter: 
An Easy Approach for Improving the Parameter-Efficiency of Adapters
}

\author{Shwai He\textsuperscript{\rm 1, \rm 4}\thanks{~~Work was done when Shwai was interning at JD Explore Academy.}\space\space\space
Liang Ding\textsuperscript{\rm 1}\thanks{~~Corresponding author}\space\space\space
Daize Dong\textsuperscript{\rm 4}\space\space\space\space
Miao Zhang\textsuperscript{\rm 2}\space\space\space
Dacheng Tao\textsuperscript{\rm 1, \rm 3}\\
    \textsuperscript{\rm 1}JD Explore Academy\\
    \textsuperscript{\rm 2}Aalborg University\space\space
    \textsuperscript{\rm 3}The university of Sydney\\
    \textsuperscript{\rm 4}University of Electronic Science and Technology of China\\
  	{\tt\small shwai.he@gmail.com},\space\space
    {\tt\small dingliang1@jd.com},\space\space
    {\tt\small dzdong2019@gmail.com},\\
    {\tt\small miaoz@cs.aau.dk},\space\space
    {\tt\small dacheng.tao@gmail.com}
}

\begin{document}
\maketitle
\begin{abstract}
Adapter Tuning, which freezes the pretrained language models (PLMs) and only fine-tunes a few extra modules, has become an appealing efficient alternative to the full model fine-tuning. Although computationally efficient, the recent adapters often increase parameters (e.g. bottleneck dimension) for matching the performance of full model fine-tuning, which we argue goes against their original intention. In this work, we re-examine the parameter-efficiency of adapters through the lens of network pruning (we name such plug-in concept as \texttt{SparseAdapter}) and find that SparseAdapter can achieve comparable or better performance than standard adapters when the sparse ratio reaches up to 80\%. Based on our findings, we introduce an easy but effective setting ``\textit{Large-Sparse}'' to improve the model capacity of adapters under the same parameter budget. Experiments on five competitive adapters upon three advanced PLMs show that with proper sparse method (e.g. SNIP) and ratio (e.g. 40\%) SparseAdapter can consistently outperform their corresponding counterpart.  Encouragingly, with the \textit{Large-Sparse} setting, we can obtain further appealing gains, even outperforming the full fine-tuning by a large margin. Our code will be released at: \url{https://github.com/Shwai-He/SparseAdapter}.
\end{abstract}

\input{Introduction}

\input{Methodology}

\input{Experiments}

\section{Conclusion}
In this work, we systematically reexamine the parameter efficiency property of adapter Tuning through the lens of network pruning. Based on our findings, we propose a plug-in strategy, i.e., SparseAdapter, for existing adapters. Our study empirically indicates the potential to make SparseAdapter (especially with the \textit{Large-Sparse} setting) a golden standard efficient transfer learning strategy for the NLP community. 

The future work includes applying our proposed SparseAdapter to more tasks (e.g. multilingual PLM based machine translation~\cite{zan2022complementarity,zan2022vega}) and benchmarks, and investigating the parameter efficiency of of other neural network models, especially for scenarios where high efficiency is required, e.g. Prompt~\cite{lester-etal-2021-power}.

\input{Limitations}

\bibliography{anthology,custom}
\bibliographystyle{acl_natbib}

\appendix

\input{Appendix}
\end{document}

%% file: Introduction.tex
\section{Introduction}
\label{:sec:Introduction}
The ``pretrain-finetune'' paradigm has become the \textit{de facto} standard for the community of natural language processing (NLP)~\cite{devlin-etal-2019-bert,liu2019roberta}. Given a pretrained language model (PLM), the conventional fine-tuning manner is tuning the entire parameters, i.e., full fine-tuning, for each downstream task~\cite{devlin-etal-2019-bert}. Considering the ever-increasing size of PLMs~\cite{NEURIPS2020_1457c0d6}, full fine-tuning has become prohibitively expensive, limiting the applicability of PLMs to a broader range of tasks.
Hence, various parameter-efficient fine-tuning approaches are explored~\cite{houlsby2019parameter,hu2021lora,Zhong2022PANDAPT}, among which \textit{Adapter Tuning}, that only tunes the extra light-weighted modules and keeps the original PLM frozen, has attached great attention. 

Despite the progress, existing adapters match the performance of full fine-tuning by increasing the bottleneck dimension~\cite{houlsby2019parameter, wang2022adamix}. This increases the overall parameters and FLOPs, violating the original intention of adapters. In this work, we turn to investigate the parameter-efficiency property (the \textit{nature} of adapters) to answer the following questions: \tcgray{\small{1}} Whether the current adapters can be further efficient? \tcgray{\small{2}} How can we increase the representation capacity of adapters within the original parameter budget? 

\begin{figure}[t]
\centering
\makeatother\def\@captype{figure}\makeatother
	\centering
    \hspace{-8pt}
	\includegraphics[width=0.47\textwidth]{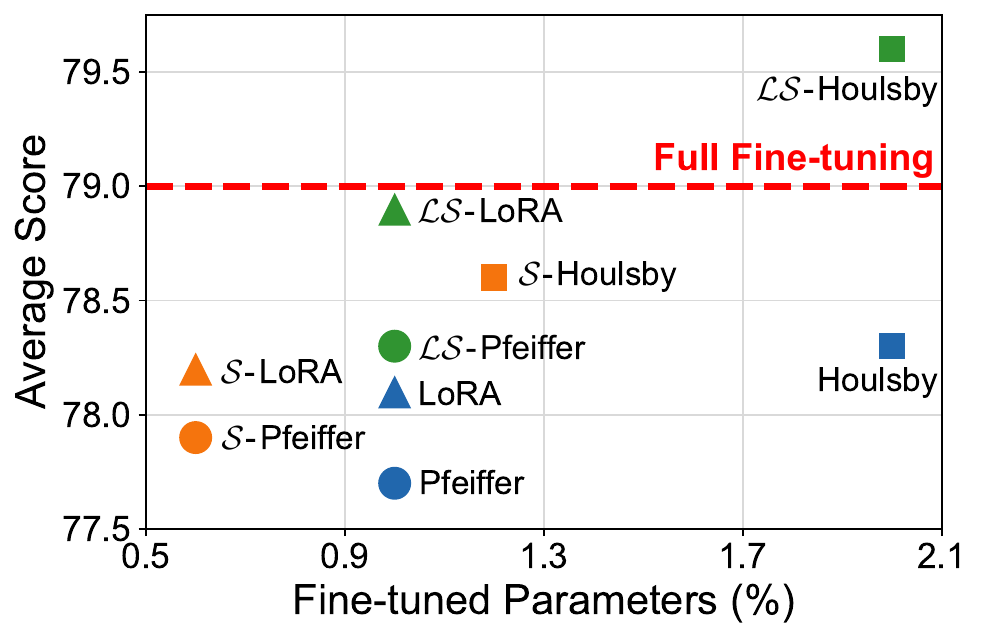}
    \caption{Performance of different parameter-efficient tuning methods on tasks from GLUE benchmark with RoBERTa-base encoder. We report the performance of Houlsby Adapters, Pfeiffer Adapters, LoRA as well as that used in our plug-in method \texttt{SparseAdapter}, where we denoted the normal sparse (in Table~\ref{main_results} and \ref{fig:variants}) as ``$\mathcal{S}$\text{-}'' and \textit{Large-Sparse} (in Table~\ref{fig:large_sparse}) as ``$\mathcal{LS}$\text{-}'' in prefix. }
    \label{fig:poster}
\end{figure}

To this end, we examine the parameter-efficiency of adapters through the lens of network pruning~\cite{mozer1989using,janowsky1989pruning}, which reduces the model size of neural networks by pruning redundant parameters and training the rest ones, therefore, improving the network efficiency. We call such pruned adapters \textbf{SparseAdapter}. Specifically, we systematically investigate five representative pruning methods in \S\ref{subsec:pruning_methods} to check at what sparse ratio can the adapters maintain the effectiveness. Note that to maintain the efficient nature of adapters, we prune all adapters at initialization such that there are no extra computational costs. We find that \tcblack{\small{1}} SparseAdapter can achieve comparable (or even better) performance than standard adapters when the sparse ratio reaches up to 80\%. Such encouraging performance could hold even using the random pruning method (See Figure~\ref{pre_experiment}) on GLUE benchmark~\cite{wang-etal-2018-glue}. Based on these insights, we introduce a frustratingly easy setting, namely \textit{Large-Sparse}, for SparseAdapter. We find that \tcblack{\small{2}} Scaling up the bottleneck dimension of SparseAdapter with a correspondingly larger sparse ratio (to ensure the same parameter budget, for example, 2$\times$ dimension scaling with 50\% sparse ratio) can effectively yield significant improvement by augmenting the model capacity. 

We validate the concept of our proposed SparseAdapter upon five advanced adapters, i.e., Houlsby~\cite{houlsby2019parameter}, Pfeiffer~\cite{pfeiffer-etal-2020-mad}, LoRA~\cite{hu2021lora}, MAM Adapter~\cite{he2022towards} and AdapterFusion~\cite{pfeiffer-etal-2021-adapterfusion}, spanning both natural language understanding (GLUE and SQuAD) and generation (XSum) benchmarks. We show that with proper sparsity, e.g. 40\%, SparseAdapter could consistently outperform their correspondingly counterpart baselines. And with our \textit{Large-Sparse} setting, SparseAdapter could even beat the full fine-tuning method significantly, e.g. 79.6 vs. 79.0 in Figure~\ref{fig:poster}.

%% file: Methodology.tex
\section{Methodology}
\label{:sec:Methodology}

\paragraph{Motivation.}
Adapters are bottleneck modules plugged in PLMs, with bottleneck dimension $r$ and model dimension $d$. In standard \textit{Adapter Tuning}, only adapter layers are trainable while the parameters of original parameters are frozen, where the number of trainable parameters determines the capacity of adapters. The common recipe to augment the capacity is to increase the bottleneck dimension, which requires more computation cost, violating the original intention of adapters. 

\begin{figure}[h]
\centering
\makeatother\def\@captype{figure}\makeatother
	\centering
        \caption{The comparison between randomly pruned adapters and standard adapters on datasets from GLUE.}
        \hspace{-8pt}
	\includegraphics[width=0.48\textwidth]{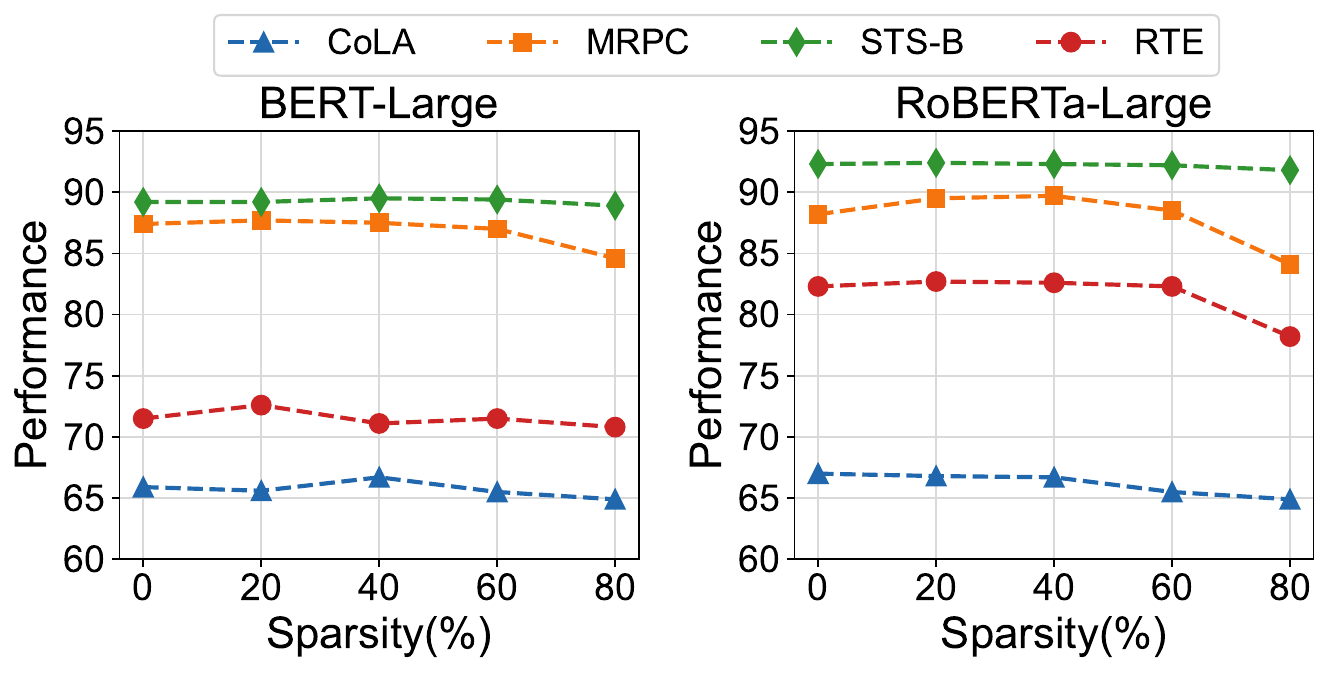}
    \vspace{-10pt}
\label{pre_experiment}
\end{figure}

To check whether augmenting adapters by increasing the parameters is an optimal choice, we decide to revisit the \textit{nature} of adapters, i.e., parameter efficiency, by pruning the redundant parameters. As shown in Figure~\ref{pre_experiment}, randomly pruned adapters can achieve comparable or even better performance than standard adapters, which indicates the existence of redundant parameters. The comparable performance could even be held under 80\% sparsity.
Such preliminary study urges us to investigate the research questions \tcgray{\small{1}} and \tcgray{\small{2}}.
We decide to approach them by systematically investigating the effects of different pruning methods. 

\begin{figure}[H]
	\centering
		\caption{Schematic comparison of (a) standard adapter and (b) our proposed SparseAdapter.}
	\begin{subfigure}[t]{0.5\textwidth}
	\includegraphics[width=\textwidth]{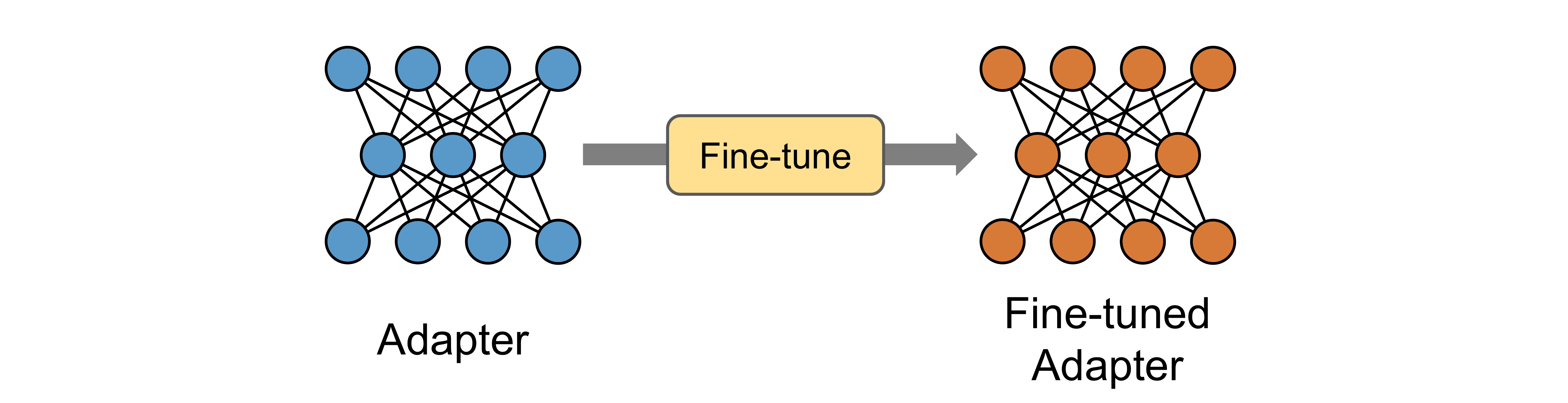}
	\caption{Standard Adapter Tuning.}
	 	 	\vspace{5pt}
 	\end{subfigure}
	\begin{subfigure}[t]{0.5\textwidth}
	\includegraphics[width=\textwidth]{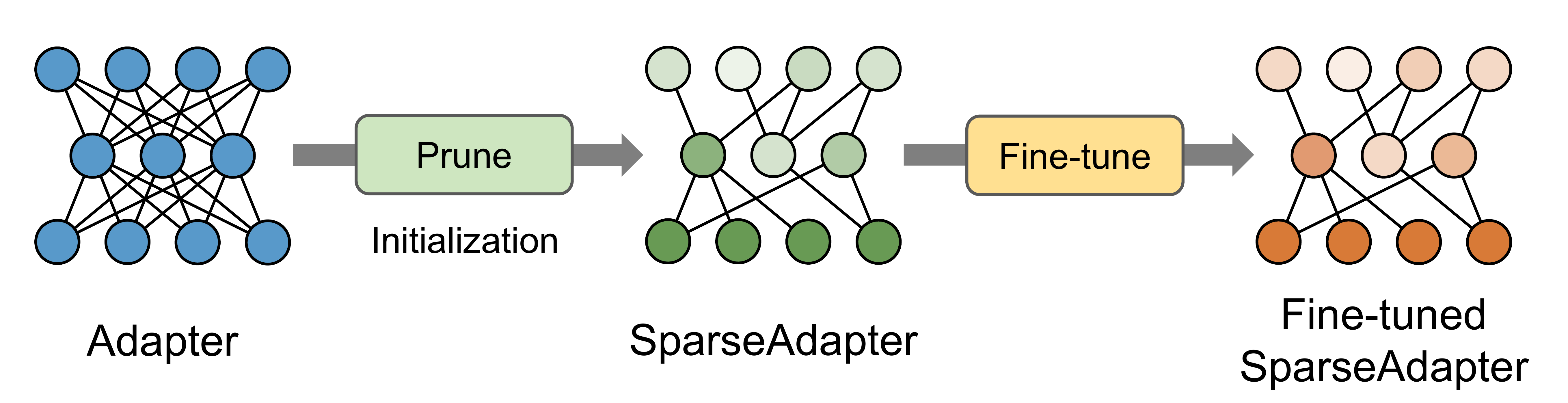}
	\caption{SparseAdapter Tuning.}
 	\end{subfigure}
	\label{fig:overview}
	\vspace{-10pt}
\end{figure} 

\subsection{Pruning Adapters at Initialization}
As is shown in Figure~\ref{fig:overview}, we intend to prune out redundant parameters and then fine-tune the SparseAdapter, instead of directly tuning all parameters (standard \textit{Adapter Tuning}). By pruning adapters at initialization, we can abandon the redundant parameters at the early stage and avoid the time-consuming iterative pruning process~\cite{frankle2018lottery}.  Specifically, considering an adapter with weights $w^l$ inserted in the layer $l \in \{1, \cdots, L\}$, parameters can be pruned by a binary mask $m^l$ as $\tilde{w}_i^l = w_i^l \odot m_i^l$, where $\tilde{w}^l_i$ denotes the pruned parameters, $w^l_i$ and $m^l_i$ denote the $i$-th element of $w^l$ and $m^l$, respectively. Given the target sparsity $s$, we assign scores $z$ to all parameters $w$ and then remove redundant parameters whose scores are below the threshold $z_s$ (the $s$-th lowest percentile of $z$). The pruning process is shown in Algorithm \ref{pruning}. 

\IncMargin{-0.5em}
\begin{algorithm}[h]
\caption{Pruning on Adapters}\label{alg:cap}
\label{pruning}
\textbf{Require:} adapter paramters $w$, sparse ratio $s$ \\
\begin{algorithmic}[1]
\State{ $w \leftarrow \operatorname{Initialization}(w)$}
\State{ $z = \operatorname{score}(w)$}
\State{ Compute the $s$-th percentile of $z$ as $z_s$}
\State{ $m \leftarrow \mathbbm{1} \left[z - {z_s} \geq 0\right]$}
\State{ $\tilde{w} \leftarrow m \odot w$}
\end{algorithmic}
\end{algorithm}

\subsection{Pruning Methods}
\label{subsec:pruning_methods}
\paragraph{Random. } Random pruning assigns a random score $z \sim \mathrm{Uniform}(0, 1)$ to each parameter and removes parameters with the lowest scores. 

\paragraph{Magnitude. } Magnitude pruning assigns each parameter with its magnitude  $z = | w| $ as its score and removes parameters with the lowest scores. Magnitude pruning is a standard way to prune during (or after) training \cite{janowsky1989pruning, han2015learning}. Here we follow \newcite{frankle2020pruning} to employ magnitude pruning at the initialization stage. 

\paragraph{Erdős-Rényi (ER). } \newcite{mocanu2018scalable, evci2020rigging} specify each layer with a random topology in which larger layers are allocated with higher sparsity than smaller layers. The layer-wise sparsity is scaled proportional to $1 - \frac{n_{in} + n_{out}}{n_{in} \cdot n_{out}}$, where $n_{in}$ and $n_{out}$ refers to the number of input and output neurons, respectively.  

\paragraph{SNIP.} \newcite{lee2018snip} compute the gradients $g_l$ for each layer with sampled mini-batch of training data, assign scores $z_l = -w_l \odot g_l$, and remove the weights with the highest scores in one iteration. The method prunes the weights with the lowest ``effect on the loss'' (either positive or negative). 

\paragraph{GraSP.} \newcite{wang2020picking} compute the Hessian-gradient product $h_l$ for each layer, issue scores $z_l = - w_l \odot h_l$, and remove the weights with the highest scores in one iteration. The method removes weights that ``reduce gradient flow'' while preserving weights that ``increase gradient flow''.

%% file: Experiments.tex
\section{Experiments}
\label{sec:Experiments}
\paragraph{Setup.}
Experiments were conducted on three widely-used benchmarks, spanning understanding and generation tasks: (1) GLUE~\cite{wang-etal-2018-glue}, containing understanding tasks like natural language inference, sentiment analysis, and sentence similarity evaluation; (2) XSum~\cite{narayan-etal-2018-dont}, a summarization dataset where the models are required to generate a short summary for a given article; (3) SQuAD v1.1~\cite{rajpurkar-etal-2016-squad}, a pair-wise dataset for questions and Wikipedia paragraphs where models select the answer span to the question from the paragraph. 

We use Adam~\cite{kingma2014adam} as the optimizer with $\beta_1$, $\beta_2$ = 0.9, 0.98. For regularization, we set the weight decay as 0.1 and grid-search the learning rate from \{1e-5, 2e-5, 5e-5, 1e-4, 2e-4\}, where we warm up the learning rate in the first 10\% steps (of the total training steps). For different data scales, we grid-search the training epoch and batch size from \{5, 10, 15, 20\}, and \{8, 16, 32, 64\}, respectively. The maximum length is 512 for GLUE and 384 for SQuAD. For XSum, we set the
max length of source articles to be 512 and the max length of the target summary to be 128. For the GLUE benchmark, we follow previous works~\cite{phang2018sentence, lee2019mixout, dodge2020fine} to fine-tune the pretrained language models, e.g. BERT~\cite{devlin-etal-2019-bert} and RoBERTa~\cite{liu2019roberta}, on the downstream training set and report results on the dev set using the last checkpoint. For the other tasks, we report the test results.

\begin{table*}[htbp]
\centering
\caption{\textbf{Experimental results of different SparseAdapters} on GLUE benchmark, where we perform pruning with the same sparsity ratio $40\%$ for a fair comparison. CoLA is evaluated using Matthew’s correlation. STS-B is evaluated using Pearson's correlation coefficient. MRPC and RTE are evaluated using accuracy. Average scores on all tasks are \underline{underlined}. The best results are \textbf{bold}. We report the results of full fine-tuning ``Fine-Tune'' as reference. }
\resizebox{\linewidth}{!}{
\begin{tabular}{lcccccccccccc}
    \toprule
    \multirow{2}{*}{\bf Method} & \multirow{1}{*}{\#Param.} & \multicolumn{5}{c}{BERT} & 
    \multicolumn{5}{c}{RoBERTa} \\
    \cmidrule(lr){3-7} \cmidrule(lr){8-12}
    &  (Trained) & \makecell{CoLA} & \makecell{MRPC} & \makecell{STS-B} & \makecell{RTE} & \underline{Avg.} 
    & \makecell{CoLA} & \makecell{MRPC} & \makecell{STS-B} & \makecell{RTE} & \underline{Avg.} \\
    \midrule
    Fine-Tune & 100\%
    & 59.4 & 83.1 & 87.2 & 68.3 & \underline{74.5} & 61.8 & 88.0 & 90.8 & 75.2 & \underline{79.0} \\
    \midrule
    Adapter & 2.0\%
    & 59.1 & 82.1 & 86.6 & 66.5 & \underline{73.6} & 61.3 & 87.4 & 90.4 & 74.1 & \underline{78.3} \\
    \hdashline
    ~~w/ Rand. & \multirow{5}{*}{1.2\%} & 58.4 & \bf 82.9 & 86.7 & 66.8 & \underline{73.7} & 61.0 & 87.5 & 90.5 & 73.2 & \underline{78.1} \\
    ~~w/ Mag.  & & 58.2 & 82.8 & 86.7 & 66.3 & \underline{73.2} & 60.6 & 87.0 & 90.6 & 73.3 & \underline{77.9} \\
    ~~w/ ER  & & 58.6 & 82.2 & 86.8 & 67.0 & \underline{73.7} & 60.9 & 87.2 & 90.2 & 73.6 & \underline{78.0} \\
    ~~w/ SNIP & & \bf 59.4 & 82.3 & \bf 87.0 & \bf 68.2 & \bf \underline{74.2} & \bf 61.4 & \bf 87.6 & 90.3 & \bf 75.0 & \bf \underline{78.6} \\
    ~~w/ GraSP & & 59.0 & 82.7 & 86.9 & 67.2 & \underline{74.0} & 61.2 & 87.1 & \bf 90.7 & 74.4 & \underline{78.4} \\
    \bottomrule
    \end{tabular}}
\label{main_results}
    \vspace{-5pt}
\end{table*}


\subsection{Results}
\paragraph{SparseAdapters with Different Pruning Methods.}
In Table~\ref{main_results}, we carefully compare SparseAdapters (with aforementioned pruning methods: ``Rand.'', ``Mag.'', ``ER'', ``SNIP'', ``GraSP'') to the standard adapter~\cite{houlsby2019parameter} (``Adapter'') on GLUE benchmark for two backbone pretrained language models BERT \cite{devlin-etal-2019-bert} and RoBERTa \cite{liu2019roberta}, where we set the bottleneck dimension to 64 for all adapter layers.  
As shown in Table~\ref{main_results}, all SparseAdapters achieve comparable or even better performance compared to Houlsby Adapter~\cite{houlsby2019parameter} with lower computational overhead. 
Notably, SNIP~\cite{lee2018snip} based SparseAdapter could achieve up to 0.6\% average improvement compared to standard adapter and nearly reach the performance of full fine-tuning, which is therefore left as the default setting in the following experiments.

\begin{table}[ht]
    \centering
    \caption{\textbf{Effect on different sparse ratios and different tasks.}
    Xsum and SQuAD are evaluated with ROUGE-2 and F1 score, respectively. We denote SparseAdapter with their sparse ratios. }
    \resizebox{\columnwidth}{!}{
    \setlength{\tabcolsep}{2pt}
    \begin{tabular}{lcccccc}
    \toprule
    \multirow{2}{*}{\bf Method} & \multicolumn{2}{c}{GLUE}& \multicolumn{2}{c}{XSum} & \multicolumn{2}{c}{SQuAD}\\
    \cmidrule(lr){2-3}
    \cmidrule(lr){4-5}
    \cmidrule(lr){6-7}
    & ~\#Para. & Avg. & ~\#Para. & R2~ & ~\#Para. & F1~\\
    \midrule
    Fine-Tune & 100\% & 79.0 & 100\% & 21.9 & 100\% & 87.8 \\
    \midrule
    Adapter & 2.0\% & 78.3 & 4.5\% & 21.6 & 8.8\% & 87.4 \\
    \hdashline
    ~$s=0.2$ & 1.6\% & \bf 78.7 & 3.6\% & 21.6 & 7.0\% & 87.5 \\
    ~$s=0.4$ 
    & 1.2\% & 78.6 & 2.7\% & \bf 21.8 & 5.3\% & \bf 87.7 \\
    ~$s=0.6$ & 0.8\% & 78.2 & 1.8\% & 21.5 & 3.5\% & 87.4 \\
    ~$s=0.8$ & 0.4\% & 77.9 & 0.9\% & 21.3 & 1.8\% & 87.0 \\
    \bottomrule
    \end{tabular}}
\label{tab:tasks}
\vspace{-5pt}
\end{table}

\paragraph{Effect on Different Downstream Tasks.}

Utilizing the proper sparse method, i.e., SNIP with 40\% sparse ratio, we validate SparseAdapter on more downstream tasks, including GLUE, XSum, and SQuAD in Table~\ref{tab:tasks}. We use RoBERTa-base \cite{liu2019roberta} for GLUE \cite{wang-etal-2018-glue}, BART-large \cite{lewis-etal-2020-bart}  for Xsum \cite{narayan-etal-2018-dont} and BERT-base \cite{devlin-etal-2019-bert} for SQuAD v1.1 \cite{rajpurkar-etal-2016-squad}. For XSum and SQuAD, The bottleneck dimension is set to 512 and 256 respectively to match the performance of full fine-tuning. Clearly, SparseAdapter outperforms the standard adapters in three tasks, showing the universality of SparseAdapter.

\paragraph{Effect on Different Sparse Ratios.} 
In Table \ref{tab:tasks}, we investigate the effect of different sparse ratios for SparseAdapter \cite{pfeiffer-etal-2021-adapterfusion}. We use BERT-base \cite{devlin-etal-2019-bert} and RoBERTa-base \cite{liu2019roberta} as backbones. SparseAdapters outperform the standard adapters when $s \leq 40 \%$ and maintained stable performance while increasing the sparse ratio. Considering the trade-off between performance and parameters, we set $40\%$ as the default sparse ratio in our work. 

\begin{table*}[ht]
\centering
\caption{{\bf Experimental results of scaling the bottleneck dimension.}  (2$\times$, 3$\times$, 4$\times$) {\bf of SparseAdapters using the same amount of parameters}, coined as \textbf{\textit{ Large-Sparse}} setting (``$\mathcal{LS}\text{-}$'' in the prefix), on GLUE benchmark. We correspondingly increase the sparsity to ensure the same number of parameters for SparseAdapters with larger bottleneck dimensions. }
\resizebox{\linewidth}{!}{
\begin{tabular}{lcccccccccccc}
    \toprule
    \multirow{2}{*}{\bf Method} 
    & 
    \multicolumn{2}{c}{Setting} & \multicolumn{5}{c}{BERT} &  \multicolumn{5}{c}{RoBERTa} \\
     \cmidrule(lr){2-3} \cmidrule(lr){4-8} \cmidrule(lr){9-13}
    & 
    $r$ & $s$ & \makecell{CoLA}  & \makecell{MRPC} & \makecell{STS-B} & \makecell{RTE} &  \underline{Avg.} & 
    \makecell{CoLA} &  \makecell{MRPC} & \makecell{STS-B} & \makecell{RTE} &  \underline{Avg.} \\
    \midrule
    Adapter & 
    $64$ & $0\%$ & 59.1 & 82.1 & 86.6 & 66.5 & \underline{73.6} & 61.3 & 87.4 & 90.4 & 74.1 & \underline{78.3} \\
    \hdashline
    \multirow{3}{*}{$\mathcal{LS}$\text{-}Adapter
    } &
    $128$ & $50\%$ & 59.9 & 82.3 & 87.6 & 67.5 & \underline{74.3} & 61.7 & 88.2 & 90.3 & 75.5 & \underline{78.9} \\
    & 
    $192$ & $67\%$ & 60.1 & 82.7 & 87.7 & 67.7 & \underline{74.6} & 61.8 & 88.7 & 90.4 & 75.3 & \underline{79.1} \\
    & 
    $256$ & $75\%$ & \bf 60.6 & \bf 83.3 & \bf 88.2 & \bf 68.2 & \underline{\bf 75.1} & \bf 62.1 & \bf 89.5 & \bf 90.5 & \bf 76.2 & \underline{\bf 79.6} \\
    \bottomrule
    \end{tabular}}
\vspace{-5pt}
\label{fig:large_sparse}
\end{table*}

\paragraph{Effect on Different Adapter Variants.}
\begin{table}[ht]
    \centering
    \caption{\textbf{Effects on other different adapter variants}. ``$\mathcal{S}$-'' means equipped with our SparseAdatper.}
    \resizebox{\columnwidth}{!}{
    \setlength{\tabcolsep}{2pt}
    \begin{tabular}{lccccc}
    \toprule
     \bf Method~ & CoLA~ & MRPC~ & STS-B~ & ~RTE~ & ~~\underline{Avg.}~ \\
     \midrule
    Pfeiffer~ 
    & \bf 61.2 & 85.8 & 89.2 & 74.7 & \underline{77.7} \\
    $\mathcal{S}$\text{-}Pfeiffer~ 
    & 61.1 & \bf 86.0 & \bf 89.3 & \bf 75.2 & \bf \underline{77.9} \\
     \midrule
    LoRA~
    & 62.0 & 87.5 & 88.5 & 74.5 & \underline{78.1} \\
    $\mathcal{S}$\text{-}LoRA~ 
    & \bf 62.1 & \bf 87.7 & \bf 88.8 & \bf 74.6 & \bf \underline{78.2}\\
     \midrule
    MAM~
    & 61.3 & 86.5 & 89.7 & \bf 74.6 & \underline{78.0} \\
    $\mathcal{S}$\text{-}MAM~ 
    & \bf 61.5 & \bf 87.6 & \bf 89.8 & 74.3 & \bf \underline{78.3} \\
    \midrule
    AF~
    & 63.1 & 89.7 & \bf 90.9 & 76.0 & 79.9 \\
    $\mathcal{S}$\text{-}AF~
    & \bf 63.3 & \bf 90.0 & 90.8 & \bf 76.4 & \bf \underline{80.1} \\
    \bottomrule
    \end{tabular}}
\label{fig:variants}
\end{table}

Since SparseAdapter can be plugged into any adapter variants, we further validate its effectiveness on other four variants besides Houlsby Apdaters in the above experiments, including
Pfeiffer~\cite{pfeiffer2020AdapterHub}, LoRA~\cite{hu2021lora}, Mix-And-Match Adapters (``MAM'')~\cite{he2022towards}, and AdapterFusion (``AF'')~\cite{pfeiffer-etal-2021-adapterfusion}. We choose RoBERTa-base \cite{liu2019roberta} as the backbone. Following previous experiments on GLUE benchmark for MAM Adapters~\cite{he2022towards}, we divide the trainable parameters equally into adapters in feed-forward layers and Prefix-Tuning \cite{li-liang-2021-prefix} in attention layers. Our SparseAdpater could consistently improve the accuracy with $40\%$ fewer training parameters, showing the generalization of our plug-in method. 
Experimental results are listed in Table~\ref{fig:variants}.
\label{sec:training_analysis}

\paragraph{Augmenting SparseAdapter with \textit{Large-Sparse} Setting.}
One strength of SparseAdapter is the potential to exploit large adapter (with a correspondingly large sparse ratio) to augment the adapter capacity under the same parameter budget, namely \textit{Large-Sparse} setting.
To validate our claim, we scale the bottleneck dimension by \{2$\times$, 3$\times$, 4$\times$\} with correspondingly \{50\%, 67\%, 75\%\} sparse ratios. As shown in Table~\ref{fig:large_sparse}, while maintaining the same amount of parameters, with bottleneck dimension increases, \textit{Large-Sparse} could consistently gain better performance, achieving up to +1.3\% and +0.6\% average improvements against the standard adapter and full fine-tuning, respectively.

\begin{figure}
\centering
\makeatother\def\@captype{figure}\makeatother
	\centering
    \caption{The comparison between SparseAdapters with \textit{Large-Sparse} setting and standard adapters. }
	\includegraphics[width=0.48\textwidth]{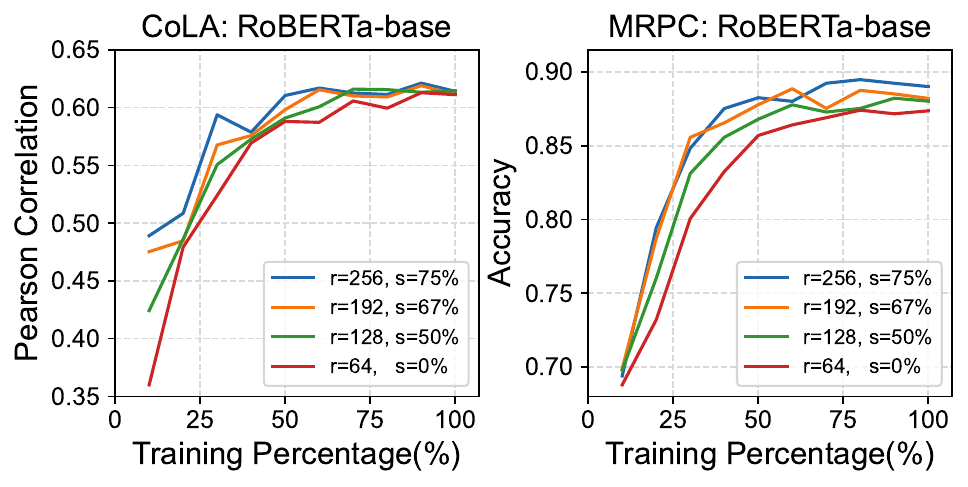}
	\vspace{-30pt}
\label{fig:training_analysis}
\end{figure}

Besides the encouraging performance, we compare SparseAdapters with \textit{Large-Sparse} setting to standard adapters on the training convergence speed in Figure \ref{fig:training_analysis}. SparseAdapters maintain a performance advantage at the same training percentage and converge at least $25\%$ ahead in the training process. For both tasks, \textit{Large-Sparse} setting contributes to a faster convergence rate and higher performance. 

%% file: Limitations.tex
\section*{Acknowledgements}
We are grateful to the anonymous EMNLP reviewers and the area chair for their insightful comments and suggestions.

\section*{Limitations}
\label{:sec:Limitations}
Despite the progress we made, there still exist limitations in our work. On the one hand, we only investigated some classic pruning methods and found that SNIP \cite{lee2018snip} performs the best in selected criteria. However, there may exist other advanced pruning methods that can further improve the performance, which deserves exploration in future work. On the other hand, since we only consider BERT, RoBERTa, and Bart in limited tasks, it would be valuable to consider other architecture families (e.g. XLNET \cite{yang2019xlnet}, ELECTRA \cite{clark2020electra}) and tasks (e.g. machine translation). 